\documentclass[10pt]{article}
 \usepackage{makeidx,epsfig,lscape}
 \usepackage{color,colortbl}
 \usepackage{fancyhdr}
 \usepackage{xcolor,pict2e}

 \renewcommand{\headrulewidth}{0pt}
 \renewcommand{\footrulewidth}{0.5pt}

 \definecolor{myaqua}{rgb}{0.0,0.5,0.55}
 \definecolor{lightaqua}{rgb}{0.75,0.95,0.95}

\usepackage{enumitem}
\usepackage{color}
\usepackage{pifont}

\usepackage{caption}

\def\lin#1#2{\textcolor[rgb]{0.6,0.6,0.6}{\vspace*{#1mm} \hrule
   height 3 pt \vspace*{#2mm}}}
\def\bt{\begin{tabular}}
\def\et{\end{tabular}}

\def\1{{\bf 1}}

 \def\sectionn#1{\refstepcounter{section}{\color{myaqua}

 \vskip 6mm

 \noindent\Large\bf\thesection. #1}

 \vskip 3mm}

\begin{document}

 \vskip 12mm

{ 

{\noindent{\huge\bf\color{myaqua}
    An Analysis of Tournament Structure
  }
  }
%
\\[6mm]
{\large\bf Nhien Pham Hoang Bao, Hiroyuki Iida}}
\\[2mm]
{ 
 Japan Advanced Institute of Science and Technology,
 1-1 Asahidai, Nomi, Japan 923-1292\\
Email:
{\color{blue}{\underline{\smash{phbnhien@jaist.ac.jp}}}},
{\color{blue}{\underline{\smash{iida@jaist.ac.jp}}}}

\lin{5}{7}

 { 
 {\noindent{\large\bf\color{myaqua} Abstract}{\bf \\[3mm]
 \textup{
 This paper explores a novel way for analyzing the tournament structures to find a best suitable one for the tournament under consideration. It concerns about three aspects such as tournament conducting cost, competitiveness development and ranking precision. It then proposes a new method using progress tree to detect potential throwaway matches. The analysis performed using the proposed method reveals the strengths and weaknesses of tournament structures. As a conclusion, single elimination is best if we want to qualify one winner only, all matches conducted are exciting in term of competitiveness. Double elimination with proper seeding system is a better choice if we want to qualify more winners. A reasonable number of extra matches need to be conducted in exchange of being able to qualify top four winners. Round-robin gives reliable ranking precision for all participants. However, its conduction cost is very high, and it fails to maintain competitiveness development.
 }}}
 \\[4mm]
 {\noindent{\large\bf\color{myaqua} Keywords}{\bf \\[3mm]
 tournament structure; competitiveness development; stability progressing; ranking precision
}

 \fancyfoot[L]{{\noindent{\color{myaqua}{\bf How to cite this
 paper:}} 
 Nh. Pham and H. Iida (2016)
 An analysis of tournament structure.
 ***********,*,***-***}}

\lin{3}{1}

\sectionn{Introduction}

{ \fontfamily{times}\selectfont
 \noindent 
 Competitive gaming does not just attract professional players only, but also many spectators who are interested in the game as well. Tournament is a competitive system. It provides some prizes as objectives for participants to compete with each other. It is often used as a formal method to conduct official gaming event, to gather professional players, as well as to attract a large number of spectators. Such large scale events usually receive sponsorship from various companies and organizations. Therefore, it is necessary to be carefully prepared and conducted to be able to avoid disappointments from any party.

\renewcommand{\headrulewidth}{0.5pt}
\renewcommand{\footrulewidth}{0pt}

 \pagestyle{fancy}
 \fancyfoot{}
 \fancyhead{} 
 \fancyhf{}
 \fancyfoot[C]{\leavevmode
 \put(0,0){\color{lightaqua}\circle*{34}}
 \put(0,0){\color{myaqua}\circle{34}}
 \put(-2.5,-3){\color{myaqua}\thepage}}

 There are three main concerns in tournament systems:
 \begin{enumerate}
 	\item Firstly, conduction cost (CC). That is, the number of matches required to conduct.
 	\item Secondly, competitiveness development (CD). That is, to minimize the throwaway matches in which participants are not so motivated to play their best. Regarding competitive gaming, some researches observe that uneven teams tend to make a reduced interest from viewers \cite{pp:Sanderson,pp:Schmidt,pp:Szymanski}. However, the structure of the tournament may have great effect on motivation of the participants. It is important to plan the matches carefully, giving the participants reasons to do their best in the game.
 	\item And finally, ranking precision (RP). That is, to make sure the ranking results of the tournament is convincing and reliable. Only that would be able to prove the prize winners are really worthy.
 \end{enumerate}
 Regarding the maintaining competitiveness issue, to the best of our knowledge, there is currently no study of any method to perform this work. Therefore, we propose a new method to analyze this matter.
 
The structure of the paper is as follows.
Section~\ref{186} presents our method for analyzing tournament structures with a focus on competitiveness development and ranking precision. 
Section~\ref{244} shows an analysis of tournament structures including single-elimination, double-elimination and round-robin style.
Section~\ref{451} discusses the analyzing results and evaluation.
Finally, concluding remarks are given in Section~\ref{471}

\section{Analyzing Method}
\label{186}
This section presents two important aspects for analyzing the quality of tournament structures: competitiveness development and ranking precision.

\subsection{Competitiveness Development}
A competitive match means that the two participants are motivated to compete over the winning outcome. Usually, the desire to win is normal. But sometimes, the benefit of winning could be insignificant which causes the participants to not yearn for a win. The motivation of a participant consists of many factors, but we restrict ourselves to the tournament structure in this study. We introduce "progress tree" to demonstrate the perspective of the participants on a tournament to analyze their motivation development throughout the tournament.

The progress tree is constructed based on the graphical model approach \cite{Sucar2015}. A participant's state before or after playing a match is considered as a node. The state in which the participant no longer plays any match is a leaf node. We show, in Figure~\ref{fig:SEBR}, an example of single elimination tournament for 8 participants, and Figure~\ref{fig:SEDT} shows its progress tree.
\begin{figure}[h]
	\centering
	\includegraphics[width=0.35\columnwidth]{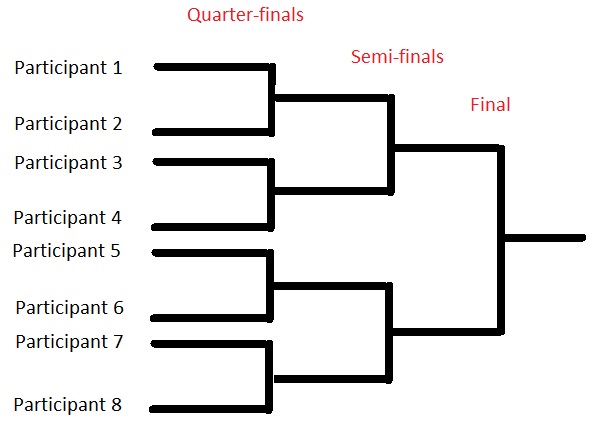}
	\caption{Single elimination tournament for 8 participants}
	\label{fig:SEBR}
\end{figure}

\begin{figure}[h]
	\centering
	\includegraphics[width=0.75\columnwidth]{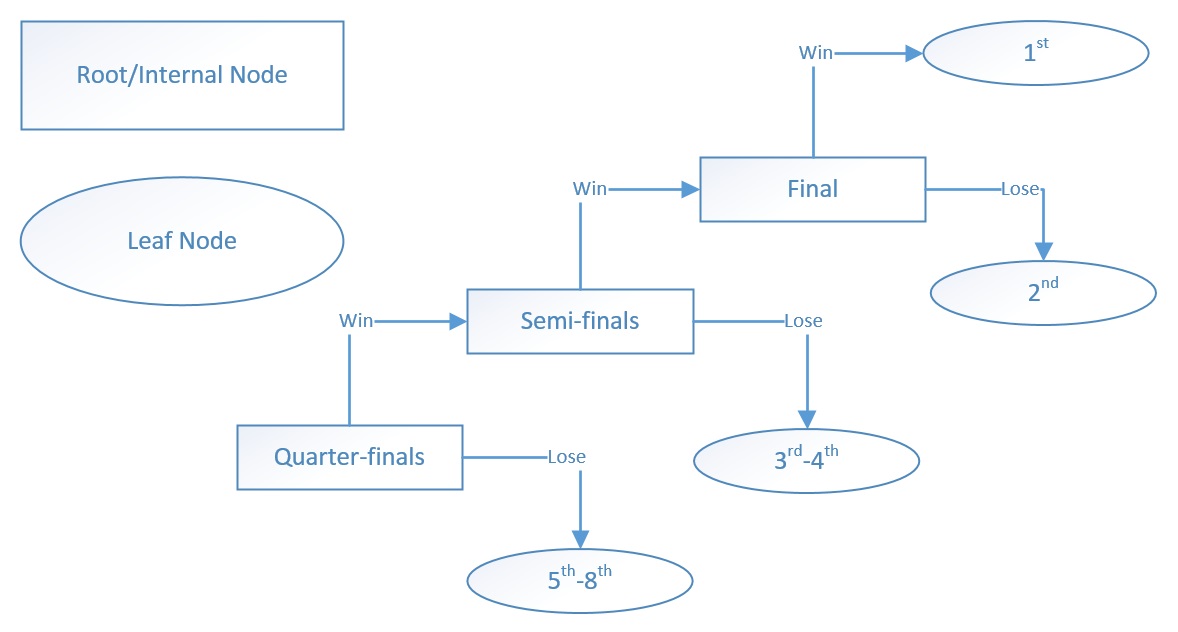}
	\caption{Progress tree of single elimination tournament for 8 participants}
	\label{fig:SEDT}
\end{figure}

While "competitive" means having an objective for which participants have to compete with one another, it is common to have more than one prize as objectives in a tournament. Thus, it is necessary to have the prizes comparable in order to have the consistency in competitiveness. For example, if we have a spoon as the first prize, and a pair chopsticks as the second prize. Each participant may have different evaluation on these prizes. Hence, it is possible to have a participant trying to lose on purpose in the final to be able to get the pair of chopsticks. This is also the reason why most grand tournaments use money for prizes instead of objects, since the amounts of money are comparable, the consistency between the prizes is ensured.

With the consistent prizes in the same unit, we would be able to evaluate nodes in the progress tree. Since we are only considering the structure of the tournament, we evaluate a node as the average value of its direct child nodes. For example, in Figure~\ref{fig:SEDT} tournament, let $x_1$, $x_2$, $x_3$ and $x_4$ be 1st place prize, 2nd prize, 3rd-4th prize and 5th-8th prize. Then we have $x1 \geq x2 \geq x3 \geq x4$. Table~\ref{tab:SEEV} shows the evaluation of other nodes. We call this value "stability" value.
\begin{table}[h]
	\caption{Evaluation of the nodes in figure \ref{fig:SEDT}}
	\label{tab:SEEV}
	\centering
	\begin{tabular}{l c}
		\small{\textbf{Node}} & \small{\textbf{Value}}\\
		\hline\\
	Final ($v$)             & $\frac{x_1+x_2}{2}$  \\\\
	Semi-final ($v_1$)      & $\frac{v + x_3}{2}$  \\\\
	Quarter-final ($v_2$)   & $\frac{v_1 + x_4}{2}$
	\end{tabular}
\end{table}

With progress tree and having the nodes evaluated, we see that there are two concerns regarding competitiveness or motivation development:
\begin{description}[align=left]
	\item [Stability progressing] 
	For every node, it is preferable to have the value of the winning outcome larger than the losing outcome. This ensures that the winning outcome has more benefits and is more attractive to the participant.
	\item [Possibility of results] 
	Since the prizes serve as an objective to maintain the competitiveness, the case where a prize is no longer able to achieve also means that a competitive objective is lost. However, in a tournament, to achieve a prize means to give up other prizes (one cannot get the first prize and second prize together). Therefore, it is favorable to have the prizes dropping out eventually in the manner of least-valuable first.
\end{description}

\subsection{Ranking Precision}
Being a competitive system, the outcome of the tournament should avoid any complains about its ranking results. We consider tournament as a comparison/sorting problem, and each match is a comparison between two participants.
We assume that there is a game where a stronger player always wins against a weaker one, and we have all participants with different strengths. There are two ways to verify the ranking precision of a tournament of such game: by mathematical proof, or by running all possibilities. In this paper we simply use the permutation to experiment all outcome possibilities. Then, we can see if the results are precise or not.
However, the method of using permutation experiment is too heavy if the number of participants is too large. In this paper, we therefore conduct experiments with eight participants only.

\section{Tournament Structure Analysis}
\label{244}
We analyze various general tournament systems such as single elimination, double elimination, and round-robin. For the purpose of comparing with each other, we demonstrate the example of having eight participants in each case.

\subsection{Single Elimination}
Single elimination is a type of elimination tournaments where the loser of each bracket is immediately eliminated.

\subsubsection{Conduction Cost}
A pure single elimination system with $i$ rounds has $n = 2^i$ participants, and there will be $n-1$ matches conducted. For 8 players single elimination, there would be 7 matches with 3 rounds.

\subsubsection{Competitiveness Development}
We use the example of 8 participants single elimination, as previously shown in Figure~\ref{fig:SEBR}.
Assuming that this tournament has comparable prizes distributed in the right order, by observing Figure~\ref{fig:SEDT} and Table~\ref{tab:SEEV}, we can see that it has no issues regarding \textit{stability progressing} or \textit{possibility of results}. All wins are worth aiming for, and the ranking results are decided lowest first.

\subsubsection{Ranking Precision}
Assuming that we have 8 participants with strength numbers vary from 1 to 8, and the stronger always win the match. These participants will be given into 8 slots as shown in Figure~\ref{fig:SEBR}. Table~\ref{tab:SEExp} shows the expected precise ranking of the tournament, while Table~\ref{tab:SE} shows the actual results counting all 8! = 40320 permutations.
\begin{table}[h]
	\caption{Precise Ranking results of single elimination for 8 participants}
	\label{tab:SEExp}
	\centering
	\begin{tabular}{l c}
		\small{\textbf{Participant}} & \small{\textbf{Precise ranking}}\\
		\hline
		8  & 1st place  \\
		7  & 2nd place  \\
		6  & 3-4th place  \\
		5  & 3-4th place  \\
		4  & 5-8th place  \\
		3  & 5-8th place  \\
		2  & 5-8th place  \\
		1  & 5-8th place
	\end{tabular}
\end{table}

\begin{table}[h]
	\caption{Ranking result of single elimination for 8 participants experiment}
	\label{tab:SE}
	\centering
	\begin{tabular}{l r r r r}
		\small{\textbf{Participant}} & \small{\textbf{1st place}} & \small{\textbf{2nd place}} & \small{\textbf{3-4th place}} & \small{\textbf{5-8th place}}\\
		\hline
		1  & 0 (0\%) & 0 (0\%) & 0 (0\%) & \textcolor{red}{40320 (100\%)}  \\
		2  & 0 (0\%) & 0 (0\%) & 5760 (14\%) & \textcolor{red}{34560 (86\%)}  \\
		3  & 0 (0\%) & 0 (0\%) & 11520 (29\%) & \textcolor{red}{28800 (71\%)}  \\
		4  & 0 (0\%) & 1152 (3\%) & 16128 (40\%) & \textcolor{red}{23040 (57\%)}  \\
		5  & 0 (0\%) & 4608 (11\%) & \textcolor{red}{18432 (46\%)} & 17280 (43\%)  \\
		6  & 0 (0\%) & 11520 (29\%) & \textcolor{red}{17280 (43\%)} & 11520 (29\%)  \\
		7  & 0 (0\%) & \textcolor{red}{23040 (57\%)} & 11520 (29\%) & 5760 (14\%)  \\
		8  & \textcolor{red}{40320 (100\%)} & 0 (0\%) & 0 (0\%) & 0 (0\%)
	\end{tabular}
\end{table}

Among all ranking results, the only 100\% correct is the 1st place. This suggests that single elimination gives convincing ranking result for the first place only, any more rankings could be complained as not being precise.

\subsection{Double Elimination}
In a standard double elimination tournament, participants are divided into two minor brackets (upper bracket and lower bracket). If a participant from the loser's bracket loses a game, the participant is eliminated. If a participant from the winner's bracket loses, the participant will be moved to the loser's bracket. The last participant remaining in the lower bracket will be the last participant standing in the upper bracket as grand final. This means that before grand final, for every upper bracket's round, there would be two rounds in lower bracket.

\begin{figure}[h]
	\centering
	\includegraphics[width=0.3\columnwidth]{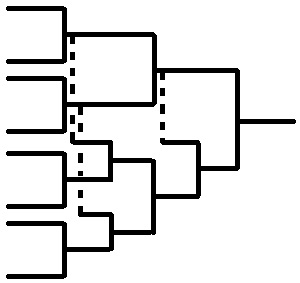}
	\caption{Double elimination tournament for 8 participants}
	\label{fig:DEBrackets}
\end{figure}

\subsubsection{Conduction Cost}
A pure double elimination system with $i$ upper rounds has $2^i$ participants, and there will be $(2^i-1)+(2^{i-1}-1)$ matches conducted. Thus, a double elimination tournament for 8 participants will have 10 matches conducted.

\subsubsection{Competitiveness Development}
We show, in Figure~\ref{fig:DEBrackets}, a double elimination tournament for 8 participants, and Figure~\ref{fig:DEDT} shows its progress tree. 
\begin{figure}
	\centering
	\includegraphics[width=0.8\columnwidth]{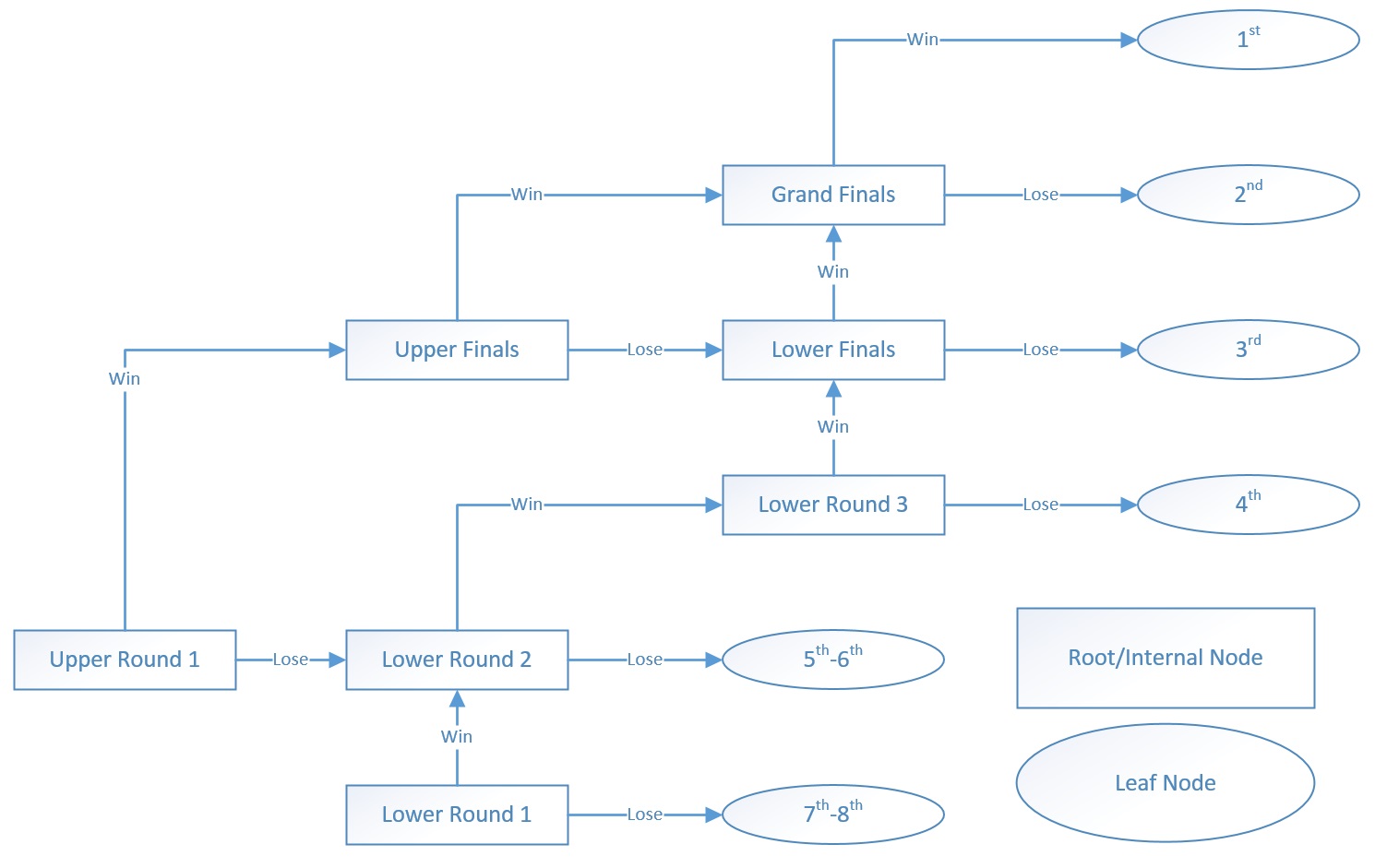}
	\caption{Progress tree of double elimination tournament for 8 participants}
	\label{fig:DEDT}
\end{figure}

Assuming that this tournament has comparable prizes distributed in the right order, by observing Figure~\ref{fig:DEDT}, we can see that it has no issues regarding \textit{stability progressing} or \textit{possibility of results}. Every win has more favorable value than its loss, and the ranking results are decided lowest first.

\subsubsection{Ranking Precision}
Assuming that we have 8 participants with strength numbers vary from 1 to 8, and the stronger always win the match. These participants will be given into 8 starting positions in Figure~\ref{fig:DEBrackets}. Table~\ref{tab:DEExp} shows the expected precise ranking of the tournament, while Table~\ref{tab:DE} shows the actual results counting all 8! = 40320 permutations.
\begin{table}[h]
	\caption{Precise Ranking result of double elimination for 8 participants}
	\label{tab:DEExp}
	\centering
	\begin{tabular}{l c}
		\small{\textbf{Participant}} & \small{\textbf{Precise ranking}}\\
		\hline
		8  & 1st place  \\
		7  & 2nd place  \\
		6  & 3rd place  \\
		5  & 4th place  \\
		4  & 5-6th place  \\
		3  & 5-6th place  \\
		2  & 7-8th place  \\
		1  & 7-8th place
	\end{tabular}
\end{table}
\begin{table}
	\caption{Ranking result of double elimination for 8 participants experiment}
	\label{tab:DE}
	\centering
	\begin{tabular}{l r r r r r r}
		\small{\textbf{Participant}} & \small{\textbf{1st place}} & \small{\textbf{2nd place}} & \small{\textbf{3rd place}} & \small{\textbf{4th place}} & \small{\textbf{5-6th place}} & \small{\textbf{7-8th place}}\\
		\hline
		1  & 0 (0\%) & 0 (0\%) & 0 (0\%) & 0 (0\%) & 20160 (50\%) & \textcolor{red}{20160 (50\%)}  \\
		2  & 0 (0\%) & 0 (0\%) & 2880 (7\%) & 0 (0\%) & 20160 (50\%) & \textcolor{red}{17280 (43\%)}  \\
		3  & 0 (0\%) & 0 (0\%) & 5760 (14\%) & 2880 (7\%) & \textcolor{red}{17280 (43\%)} & 14400 (36\%)  \\
		4  & 0 (0\%) & 576 (1\%) & 8064 (20\%) & 7488 (19\%) & \textcolor{red}{12672 (31\%)} & 11520 (27\%)  \\
		5  & 0 (0\%) & 2304 (6\%) & 9216 (23\%) & \textcolor{red}{12672 (31\%)} & 7488 (19\%) & 8640 (21\%)  \\
		6  & 0 (0\%) & 5760 (14\%) & \textcolor{red}{14400 (36\%)} & 11520 (27\%) & 2880 (7\%) & 5760 (14\%)  \\
		7  & 0 (0\%) & \textcolor{red}{31680 (79\%)} & 0 (0\%) & 5760 (14\%) & 0 (0\%) & 2880 (7\%)  \\
		8  & \textcolor{red}{40320 (100\%)} & 0 (0\%) & 0 (0\%) & 0 (0\%) & 0 (0\%) & 0 (0\%)
	\end{tabular}
\end{table}

Even with increased number of matches conducted compared to single elimination, it is still very lacking in the ranking precision. However, double elimination has two starting nodes (see  Figure~\ref{fig:DEDT}). It can be argued that double elimination is designed for a multi-stage system, or the players are distributed properly by using a rating system \cite{elo,5446231,trueskilltm-a-bayesian-skill-rating-system,trueskill-through-time-revisiting-the-history-of-chess,5283063,5671406}. Thus, it is expected to have stronger participants and weaker participants distributed (seeded) into upper bracket and lower bracket properly. We perform another experiment this way, as a seeded double elimination. There will be $4! * 4! = 576$ cases this time.

\begin{table}
	\caption{Ranking result of seeded double elimination for 8 participants experiment}
	\label{tab:DES}
	\centering
	\begin{tabular}{l r r r r r r}
		\small{\textbf{Participant}} & \small{\textbf{1st place}} & \small{\textbf{2nd place}} & \small{\textbf{3rd place}} & \small{\textbf{4th place}} & \small{\textbf{5-6th place}} & \small{\textbf{7-8th place}}\\
		\hline
		1  & 0 (0\%) & 0 (0\%) & 0 (0\%) & 0 (0\%) & 0 (0\%) & \textcolor{red}{576 (100\%)}  \\
		2  & 0 (0\%) & 0 (0\%) & 0 (0\%) & 0 (0\%) & 192 (33\%) & \textcolor{red}{384 (67\%)}  \\
		3  & 0 (0\%) & 0 (0\%) & 0 (0\%) & 0 (0\%) & \textcolor{red}{384 (67\%)} & 192 (33\%)  \\
		4  & 0 (0\%) & 0 (0\%) & 0 (0\%) & 0 (0\%) & \textcolor{red}{576 (100\%)} & 0 (0\%)  \\
		5  & 0 (0\%) & 0 (0\%) & 0 (0\%) & \textcolor{red}{576 (100\%)} & 0 (0\%) & 0 (0\%)  \\
		6  & 0 (0\%) & 0 (0\%) & \textcolor{red}{576 (100\%)} & 0 (0\%) & 0 (0\%) & 0 (0\%)  \\
		7  & 0 (0\%) & \textcolor{red}{576 (100\%)} & 0 (0\%) & 0 (0\%) & 0 (0\%) & 0 (0\%)  \\
		8  & \textcolor{red}{576 (100\%)} & 0 (0\%) & 0 (0\%) & 0 (0\%) & 0 (0\%) & 0 (0\%)  \\
	\end{tabular}
\end{table}

Table~\ref{tab:DES} shows the result of the seeded double elimination. It is precise from 1st to 4th ranking. This is quite a big improvement compared to single elimination, but with the cost of having a little more matches and having a proper seeding system.

\subsection{Round-Robin}
In round-robin tournament system, all participants have to play with each other. In other words, each participant plays every other participant once. If each participant plays all others twice, it is called double round-robin.

\subsubsection{Conducting Cost}
A pure round-robin system with $n$ participants has $\frac{n}{2}(n-1)$ matches conducted. Thus for 8 participants, there would be 28 matches.

\subsubsection{Competitiveness Development}
We show, in Figure~\ref{fig:RRDT}, a progress tree of a round-robin tournament with 8 participants. The big difference from elimination tournaments is that from the beginning, only the leaf from all losses and the leaf from all wins are known. This unstable situation makes us unable to calculate stability values of the nodes. As the tournament progresses, the unknown leaves will gradually reveal, and the stability values of the nodes would be calculated. Furthermore, unstable situations suggest that there are possibilities of not satisfying \textit{stability progressing} and \textit{possibility of results} conditions.

\begin{figure}[h]
	\centering
	\includegraphics[width=1\columnwidth]{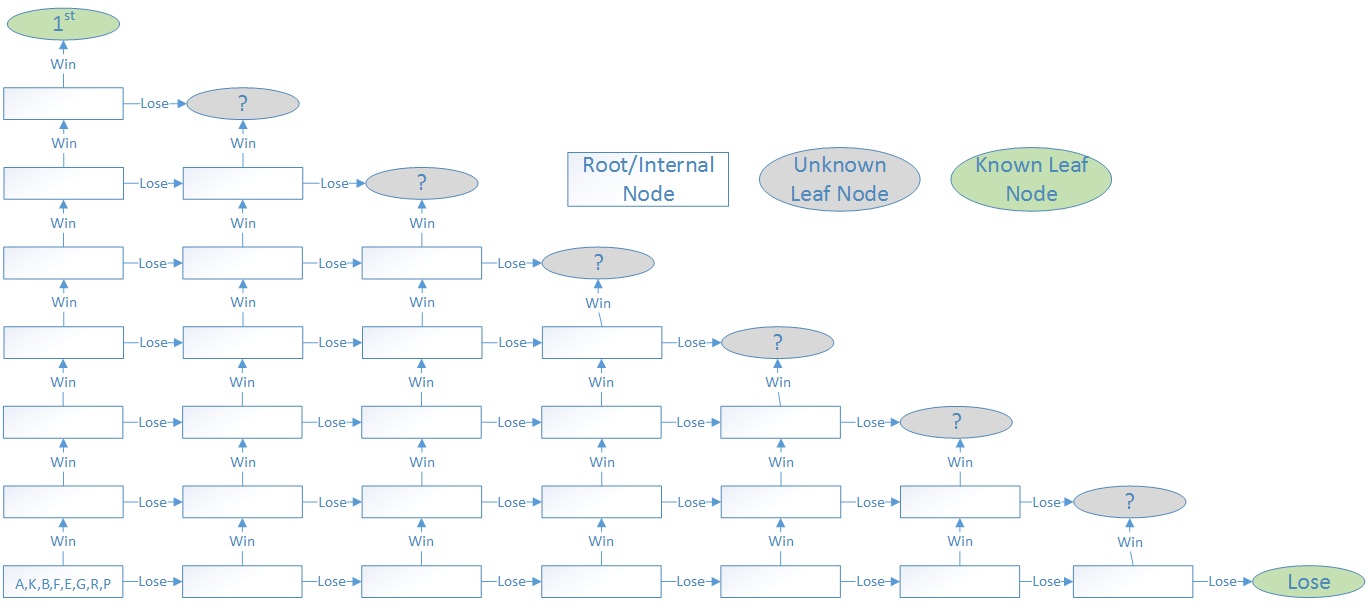}
	\caption{First round progress tree of round-robin tournament for 8 participants}
	\label{fig:RRDT}
\end{figure}

We show, in Table~\ref{tab:RRRS}, an example situation after 5 rounds, and Figure~\ref{fig:RRDT5} shows its progress tree. In this situation, if participant A wins the next match, his victory as 1st place would be fixed regardless of his last match outcome. This fails to satisfy \textit{stability progressing}. Besides, for participants B, F, E, G, R, and P, even their leaves are unknown, the possibility of 1st place is out of reach. Therefore, this situation does not satisfy \textit{possibility of results} either.

\begin{table}[h]
	\caption{An possibility of round-robin tournament with 8 participants (after 5 rounds)}
	\label{tab:RRRS}
	\centering
	\begin{tabular}{c c c}
		\small{\textbf{Participant}} & \small{\textbf{Wins}} & \small{\textbf{Losses}}\\
		\hline
		A & 5 & 0\\
		K & 3 & 2\\
		B & 2 & 3\\
		F & 2 & 3\\
		E & 2 & 3\\
		G & 2 & 3\\
		R & 2 & 3\\
		P & 2 & 3
	\end{tabular}
\end{table}
\begin{figure}
	\centering
	\includegraphics[width=1\columnwidth]{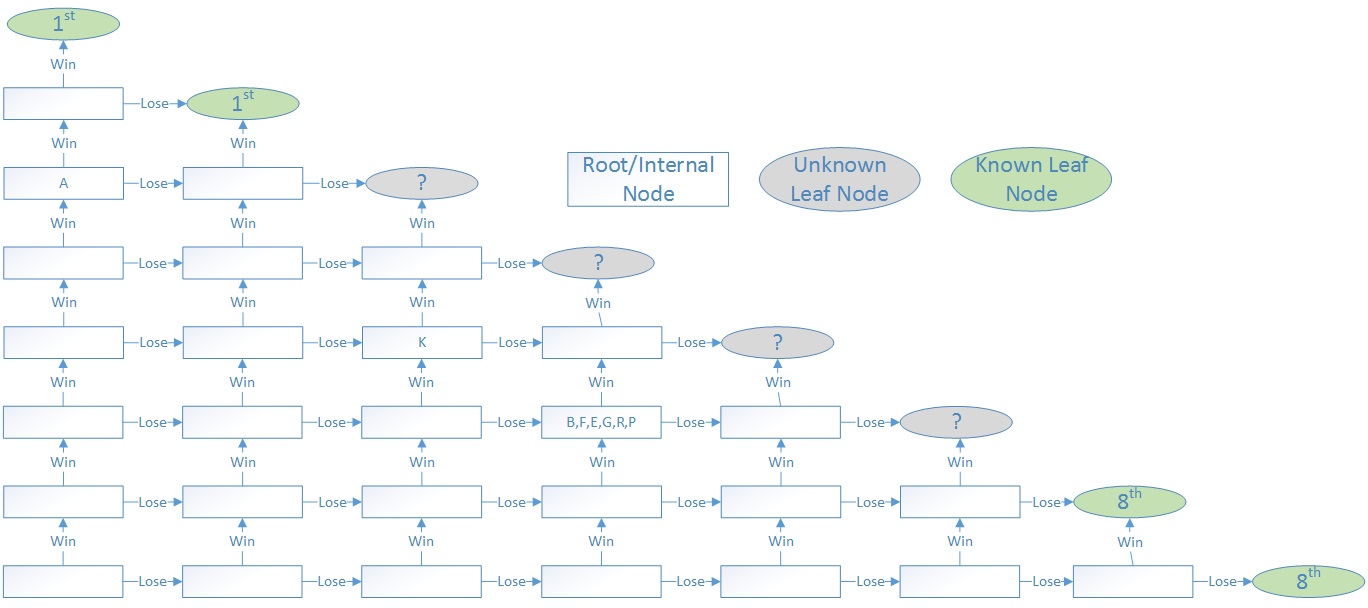}
	\caption{Progress tree for the situation in table \ref{tab:RRRS}}
	\label{fig:RRDT5}
\end{figure}

\subsubsection{Ranking Precision}
Assuming that we have 8 participants with strength numbers vary from 1 to 8, and stronger participant always win. Then, there is only one outcome as in Table~\ref{tab:RREXP}, no matter the participants are positioned.
\begin{table}[h]
	\caption{Experimental result of round-robin for 8 participants}
	\label{tab:RREXP}
	\centering
	\begin{tabular}{c c c r}
		\small{\textbf{Participant}} & \small{\textbf{Wins}} & \small{\textbf{Losses}} & \small{\textbf{Ranking}}\\
		\hline
		8 & 7 & 0 & 1st place\\
		7 & 6 & 1 & 2nd place\\
		6 & 5 & 2 & 3rd place\\
		5 & 4 & 3 & 4th place\\
		4 & 3 & 4 & 5th place\\
		3 & 2 & 5 & 6th place\\
		2 & 1 & 6 & 7th place\\
		1 & 0 & 7 & 8th place
	\end{tabular}
\end{table}
Round-robin gives a really accurate ranking in the experiment. However, the conduction cost is high, and the competitiveness development is not good.

\section{Results and Evaluation}
\label{451}
We show in Table~\ref{tab:eval} the comparison between single elimination, double elimination, and round-robin. We can see that higher conduction cost is needed to achieve better ranking precision. The result shows that the single elimination system has the lowest cost, while its competitiveness development is properly maintained. However, its ranking is only reliable on the top one winner only. Double elimination has no problem in competitiveness development either. With proper seeding system, although it has higher conduction cost, its ranking is precise on the top 4 winners. Round-robin on the other hand gives precise ranking on all participants, but may have too heavy cost which causes the lacking in competitiveness development.

\begin{table}[h]
	\caption{
	Strength and weakness of tournament structures. \newline Single-elimination (SE), Double-elimination (DE), Seeded double-elimination (DE Seeded) and Round-robin (RR) compared
	by conducting cost (CC), competitiveness development (CD) and ranking precision (RP)}
	\label{tab:eval}
	\centering
	\begin{tabular}{l r c l}
		\small{\textbf{System}} & \small{\textbf{CC}} & \small{\textbf{CD}} & \small{\textbf{RP}}\\
		\hline
		SE & Low (7) & \ding{51} & Top 1 winner only\\
		DE & Medium (10) & \ding{51} & Top 1 winner only\\
		DE (Seeded) & Medium (10) & \ding{51} & Top 4 winners\\
		RR & High (28) & \ding{55} & All
	\end{tabular}
\end{table}

\section{Concluding Remarks}
\label{471}
%


In this paper we proposed a novel way for analyzing the tournament structures to find a best suitable one for the tournament under consideration. We focused on three aspects such as tournament conducting cost, competitiveness development and ranking precision. We proposed a notion of progress tree to detect potential throwaway matches. The analysis performed using the proposed method reveals the strengths and weaknesses of tournament structures. As a conclusion, single elimination is best if we want to qualify one winner only, all matches conducted are exciting in term of competitiveness. Double elimination with proper seeding system is a better choice if we want to qualify more winners. A reasonable number of extra matches need to be conducted in exchange of being able to qualify top four winners. Round-robin gives reliable ranking precision for all participants. However, its conduction cost is very high, and it fails to maintain competitiveness development.

Our future works will investigate other tournaments system such as Swiss and its variants, eventually find out the most balanced tournament structure in term of conduction cost, competitiveness development, and ranking precision.

 {\color{myaqua}

}}

\end{document}